\documentclass[letterpaper]{article} %DO NOT CHANGE THIS
\usepackage[preprint,nonatbib]{nips_2018}  %Required
\usepackage{booktabs}
\usepackage{times}  %Required
\usepackage{helvet}  %Required
\usepackage{courier}  %Required
\usepackage{url}  %Required
\usepackage{graphicx}  %Required
\usepackage{rotating}
\usepackage{wrapfig}
\usepackage{multicol}
\usepackage{latexsym}
\usepackage{amsmath}
\usepackage{amssymb}
\usepackage{url}
\usepackage[noend]{algorithmic}
\usepackage{algorithm}
\usepackage{multirow}
\usepackage{mathtools}
\usepackage{enumitem}
\usepackage{color}
\usepackage{microtype}
\usepackage{tabularx}
\usepackage{relsize}
\usepackage{amsthm}

\newtheoremstyle{mytheoremstyle} % name
{\topsep}                    % Space above
{\topsep}                    % Space below
{\itshape}                   % Body font
{}                           % Indent amount
{\scshape}                   % Theorem head font
{.}                          % Punctuation after theorem head
{.5em}                       % Space after theorem head
{\thmname{#1}\thmnumber{ #2}\thmnote{ (#3)}}  % Theorem head spec (can be left empty, meaning ‘normal’)

\theoremstyle{mytheoremstyle}
\DeclarePairedDelimiter\ceil{\lceil}{\rceil}
\newtheorem{thm}{Theorem}

\newtheorem{prop}[thm]{Proposition}

\newcommand{\mcesp}{\textsf{MCES-P}}

\newcommand{\mcesmp}{\textsf{MCES-MP}}

\newcommand{\mcesmppac}{\textsf{MCESMP}\textsf{+PALO}}
\newcommand{\mcesposg}{\textsf{MCES-MP}}
\newcommand{\mcesfp}{\textsf{MCES-FMP}}

\newcommand{\mcesfppac}{\textsf{MCESFMP+PALO}}
\newcommand{\R}{\mathcal{R}}

\begin{document}

\title{Reinforcement Learning for Heterogeneous Teams with PALO Bounds}  % put your title here!
%\titlenote{Produces the permission block, and copyright information}

\author{Roi Ceren\\
Department of Computer Science\\
University of Georgia, Athens, GA 30602\\
\texttt{roi.ceren@gmail.com}  % put your paper number here!
\And
Prashant Doshi\\
Department of Computer Science\\
University of Georgia, Athens, GA 30602\\
\texttt{pdoshi@cs.uga.edu}
\And
Keyang He\\
Department of Computer Science\\
University of Georgia, Athens, GA 30602\\
\texttt{kh454336@uga.edu}
}

\maketitle

\begin{abstract}  % put your abstract here!
  We introduce reinforcement learning for heterogeneous teams in which
  rewards  for an  agent  are additively  factored  into local  costs,
  stimuli unique  to each agent,  and global rewards, those  shared by
  all agents  in the domain.  Motivating  domains include coordination
  of varied  robotic platforms,  which incur  different costs  for the
  same action, but share an overall goal. We present two templates for
  learning in this setting with  factored rewards: a generalization of
  Perkins'  Monte  Carlo  exploring  starts for  POMDPs  to  canonical
  MPOMDPs,  with a  single policy  mapping joint  observations of  all
  agents to  joint actions  (\mcesmp{}); and  another with  each agent
  individually  mapping   joint  observations  to  their   own  action
  (\mcesfp{}).   We use  probably approximately  local optimal  (PALO)
  bounds to  analyze sample complexity, instantiating  these templates
  to  PALO learning.   We  promote sample  efficiency  by including  a
  policy space pruning technique, and evaluate the approaches on three
  domains of heterogeneous agents  demonstrating that \mcesfp{} yields
  improved  policies  in less  samples  compared  to \mcesmp{}  and  a
  previous benchmark.
\end{abstract}

%%%%%%%%%%%%%%%%%%%%%%%%%%%%%%%%%%%%%%%%%% start of main body of paper

%\vspace{-0.1in}
\section{Introduction}
\label{sec:intro}

\begin{figure}[h]
  \vspace{-0.1in}
	\centering
	\includegraphics[width=1.35in]{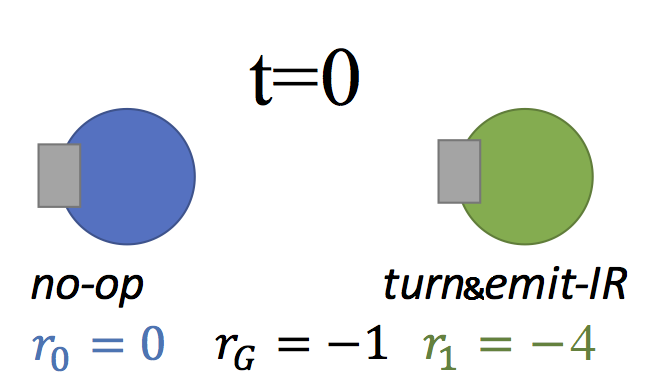} 
        \includegraphics[width=1.35in]{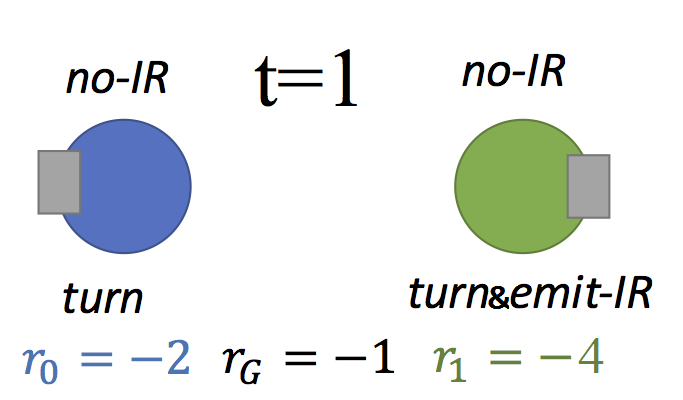}
	\includegraphics[width=1.2in]{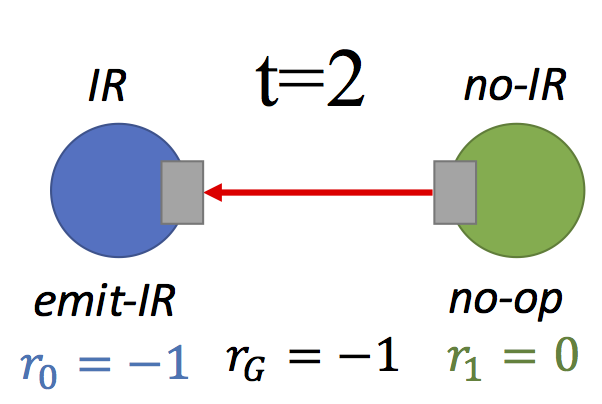} 
        \includegraphics[width=1.2in]{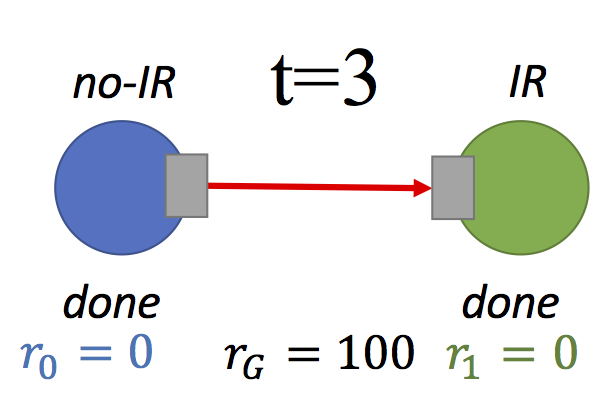}
        \vspace{-0.05in}
        \caption{Illustration of  a converged policy on  the two-robot
          alignment  problem yielding  a trajectory  where the  robots
          align. Rewards $r_0$  and $r_1$ are local  costs while $r_G$
          is the global team reward.}
\label{fig:robot}
\vspace{-0.05in}
\end{figure}

We  focus on  optimizing the  performance of  a team  of heterogeneous
agents in  partially observable environments without  knowledge of the
environmental dynamics. An illustrative example is the robot alignment
problem~\cite{Kraemer16:Multi-agent}  shown  in  Fig.~\ref{fig:robot},
where two different  platforms seek to face each other  by turning and
emitting infrared signals,  which the other detects.   Here, one robot
turns  inexpensively due  to  sophisticated  actuators incurring  less
energy cost.
%while the other expends more, which characterizes the team's heterogeneity.

We consider  settings comprised of  two or  more agents situated  in a
partially observable environment, where each agent receives a possibly
noisy   observation.   Furthermore,   each  agent   receives  both   a
\textit{local} reward based on the state and its individual action, as
well  as   a  \textit{global}  team   reward  based  on   all  agents'
actions. Factoring  rewards into local  and global components  is well
explored        in        cooperative       multiagent        decision
making~\cite{chang2004all,factoredmdp,liuopt}  offering   the  benefit
that cooperation among heterogeneous agents,  such as a team of varied
robotic platforms, may  be studied. {\em Both  observation and rewards
  of  each  agent   are  communicated  instantly  and   exactly  to  a
  centralized learner.} The joint  transition, observation, and reward
functions are unknown making this {\em model-free} RL. This form of RL
is  well suited  for  robot  learning where  motion  noise levels  are
specific and usually not known.

Perkins'~\cite{perkins}  Monte Carlo  exploring  starts for  partially
observable  Markov decision  processes (POMDP)  (labeled as  \mcesp{})
performs  online sampling  of trajectories  and explores  local policy
neighborhoods,    terminating   exploration    when   the    empirical
action-values for a  policy can no longer be improved  by performing a
different action. When leveraging  sample count requirements analogous
to  probably  approximately  correct bounds,  \mcesp{}  most  probably
arrives at an $\epsilon$-{\em locally optimal} policy where $\epsilon$
is     a    function     of     the    sample     count    and     the
probability~\cite{PALO}. \mcesp{} offers the  benefit of model-free RL
with  PALO  guarantees  in   partially  observable  state  spaces  and
single-agent contexts.

With  \mcesp{} as  the  departure  point, this  paper  makes two  main
contributions. First, it introduces two templates for model-free RL in
the heterogeneous  setting.  Our  first template is  a straightforward
extension of \mcesp{}, which models  the problem as a canonical MPOMDP
by mapping joint observation sequences  to joint actions.  We label it
as  \mcesmp{}. The  second  approach, motivated  by methods  factoring
joint-policy    search   spaces~\cite{Amato07:Optimizing,factoredmdp},
decomposes the  single large  policy in \mcesmp{}  to a  collection of
policies each  mapping joint observations to  individual actions. This
template that performs  MCES for each agent and  improves the policies
only if all  agents show an improvement is labeled  as \mcesfp{}. Both
these templates not only offer model-free RL in the context of MPOMDPs
for the  first time, they  represent first methods that  relate sample
bounds to local optimality for RL in a popular multiagent context.

Next,  we  instantiate  our  methods   with  PALO  bounds  to  provide
statistical guarantees of $\epsilon$-local optimality that relate with
sample  complexity. We  empirically demonstrate  in three  cooperative
problem domains that both approaches  arrive at good local optima with
varying domain parameters. We  illustrate the comparative advantage of
the decomposition in \mcesfp{}, providing remarkable sample complexity
savings over the \mcesmp{}  instantiation.  We exploit a parameterized
policy  search  space   pruning  introduced  previously~\cite{mcesip},
trading statistical guarantees  from PALO bounds for  reduction in the
computational burden. In comparisons with a known method, \mcesfp{} is
sample efficient and  learns much better policies, though,  as is well
known, PALO guarantees limit scalability.

Our  cooperative multiagent  setup differs  significantly from  single
agent  settings.  Furthermore,  the factored  rewards model  structure
that  is  well suited  to  robotic  settings.   While \mcesmp{}  is  a
straightforward  generalization  of  \mcesp{}, \mcesfp{}  is  a  novel
contribution. It offers much reduced sample complexity due to which it
runs faster and requires less pruning.

\vspace{-0.1in}
\section{Background on RL for POMDPs}
\label{sec:background}
\vspace{-0.1in}

Monte Carlo exploring starts for POMDPs (\mcesp{})~\cite{perkins} uses
MCES~\cite{rl},  a   flavor  of   Q-learning  that   randomly  selects
state-action  pairs to  evaluate  in the  beginning. \mcesp{}  locally
explores neighboring policies,  those that differ by  a single action,
and hill climbs  to those that have  empirically-derived higher reward
values.   It   explores  the  local   neighborhood  of  a   policy  by
transforming  the  action taken  at  a  randomly selected  observation
sequence $\vec{o}$.  After each transformation, \mcesp{}  compares the
empirical  reward  following  the  transformed  observation  sequence,
denoted by $Q_{\vec{o},a}$, against the  current policy.  For this, it
generates  a  $T$-step  sampled  trajectory  containing  observations,
actions,                          and                         rewards:
$\tau=(a^0,r^0,o^1,a^1,r^1,...o^{T-1},a^{T-1},r^{T-1})$.           Let
$\R(\tau)=\sum_{t=0}^{T-1}\gamma^t  r^t$  be  the  discounted  sum  of
rewards  for a  trajectory where  $\gamma \in  (0,1)$ is  the discount
factor.  Let  $R_{pre-\vec{o}}(\tau)$ and  $R_{post-\vec{o}}(\tau)$ be
the cumulative  rewards up to  and following the  observation sequence
$\vec{o}$  in   a  trajectory  $\tau$,  respectively.    The  expected
action-value of a policy is then:
%\begin{small}
	\begin{flalign*}
		\mathsmaller{Q_{\pi}}   \triangleq E^{\tau \sim \pi} [R(\tau)] = E^{\tau \sim \pi} [ R_{pre-\vec{o}}(\tau) + R_{post-\vec{o}}(\tau) ]
		= E^{\tau \sim \pi} [ R_{pre-\vec{o}}(\tau) ] + E^{\tau \sim \pi} [ R_{post-\vec{o}}(\tau) ] 
	\end{flalign*}
%\end{small}

        \mcesp{}  updates the  Q-value  by  $R_{post-\vec{o}}$ with  a
        depreciating  learning rate.  At each  stage, $k$  samples are
        taken of the value of a current policy and of each neighboring
        policy. It also prevents comparing policies that have not been
        sampled sufficiently.
%\mcesp{} updates the Q-value by $R_{post-\vec{o}}$ with a depreciating
%learning rate $\alpha(m, c) =  \frac{1}{c+1}$, where $m$ is the number
%of transformations  taken so far and  $c$ the count of  updates to the
%Q-function. In this instance, $m$ does not affect the learning rate. At each stage, $k$ samples are taken of the value of a current policy and of each neighboring policy. It prevents comparing policies that have not been sampled
%sufficiently by setting $\epsilon(m,p,q) = +\infty$ if $p<k$ or $q<k$,
%and  $0$ otherwise. Here,  $p$ and  $q$ are  the number  of trajectory
%samples  for  original  and  transformed policies,  respectively.   If
        If  Q-values for  the  transformation dominate  the policy  by
        $\epsilon$, it is accepted and the process repeats for the new
        policy.  Termination  is triggered when, after  $k$ samples of
        trajectories  that update  the  Q-value of  each neighbor,  no
        neighbor dominates current policy.

        An   instantiation  of   \mcesp{}  using   Greiner's  probably
        approximately  correct  local optimization~\cite{PALO}  allows
        the  inclusion  of  PAC-like  guarantees,  leveraging  a  more
        refined  $\epsilon$ and  $k$.   In addition,  a  bound on  the
        probability  of error,  $\delta$, is  introduced.  The  sample
        requirement $k_m$ and error  probability bound $\delta$ change
        with the number of transformations $m$.
%and are derived as,
\begin{small}
\begin{align}
	k_m\gets\ceil*{2\left(\frac{\Lambda}{\epsilon}\right)^2 ln\frac{2N}{\delta_m}}
  \label{eq:k_m}
\end{align}
\end{small}
where $\delta_m\gets6\delta/9.872 m^2$. Let $p$ and $q$ be the numbers
of  samples  of  the  two   policies  under  comparison,  $N$  is  the
cardinality of the set of neighboring policies. Then, error $\epsilon$
is defined as:
\begin{small}
\begin{align}
          \epsilon(m,p,q) = \left\{
            \begin{array}{lr}
              \epsilon^*(m,p) & \text{if } p=q < k_m\\%\nonumber\\
              \frac{\epsilon}{2} & \text{if } p = q = k_m\\%\nonumber\\
              +\infty & \text{otherwise}%\nonumber
            \end{array}
    \right. 
                              & \text{where }
                                \epsilon^*(m,p)=\Lambda~\sqrt{\frac{1}{2p}\ln\left(\frac{2(k_m-1)N}{\delta_m}\right)}.
\label{eq:epsilon}
\end{align}
\end{small}
As   policies   differ    by   a   single   action    only,   $N$   is
$\mathcal{O}(|A|{\frac{|\Omega|^T-1}{|\Omega|-1}})$  --- significantly
less      than       the      entire      space       of      policies
$\mathcal{O}({|A|}^{\frac{|\Omega|^T-1}{|\Omega|-1}})$;     $A$    and
$\Omega$ are the sets of actions and observations, respectively, while
$T$ is the planning horizon.
  
Let $\pi$ be a current policy,  $\pi'$ a local neighbor, 
$\Lambda(\pi,\pi')$ the range  of their possible action-values, and $R_{max}$, $R_{min}$ be  the  maximum and  minimum
possible rewards from a single step. Then,\vspace{-0.07in}
\begin{small}
\begin{align}
  \Lambda(\pi,\pi')  & \triangleq \max_{\tau} (Q_{\pi}(\tau) -
                       Q_{\pi'}(\tau)) - \min_{\tau} (Q_{\pi}(\tau) -
                       Q_{\pi'}(\tau)) \leqslant \sum\nolimits_{t =0
                       \ldots T-1} \left [(R_{max} -
                       R_{min})\right. \nonumber\\ 
& ~~~ \left . - (R_{min} - R_{max})\right ] = \sum\nolimits_{t=0\ldots T-1} 2(R_{max}- R_{min})  
                         = 2T(R_{max}- R_{min}).
                         \label{eq:lambda} 
	\end{align}
  \end{small}
  Then,                                                            let
  $\Lambda                 \triangleq                \max_{\pi,\pi'\in
    neighbor(\pi)}\Lambda(\pi,\pi')$.    The  PALO   instantiation  of
  \mcesp{} is implemented by utilizing the redefinitions of $\epsilon$
  and $k_m$.  It terminates when no policy transformation is triggered
  after  $k_m$  samples,  or  if  the  current  policy  dominates  its
  neighbors  by   $\epsilon-\epsilon^*(m,p)$  for  fewer   than  $k_m$
  samples.     Hoeffding's    inequality     and    Theorem    2    of
  Perkins~\cite{perkins}   guarantee  that   the  PALO   instantiation
  converges  to  $\epsilon$-locally  optimal policy  with  probability
  $\geq 1-\delta$.
        %guarantees with
        %probability   $1-\delta$  that   the   convergent  policy   is
        %$\epsilon$-locally optimal.

\vspace{-0.1in} 
\section{Related Work}
\label{sec:relatedworks}
\vspace{-0.1in} 
%\vspace{-0.05in}

\mcesp{}   has    been   extended   to    self-interested   multiagent
settings~\cite{mcesip}, where  opponents' policies  are fixed  but not
known to the subject agent. It  maintains beliefs over models of other
agents. The self-interested setting is orthogonal to the \textit{team}
problem studied in this paper.

In the  context of MPOMDPs, factored-value  partially observable Monte
Carlo planning~\cite{amato}  offers some scalability by  factoring the
joint  {\em  value  function}   to  exploit  structure  in  multiagent
systems. However, this is  essentially a {\em model-based} centralized
approach  that  either requires  prior  knowledge  of the  environment
dynamics or an  augmented model involving beliefs  over the transition
function             learned            via             Bayes-Adaptive
MPOMDP~\cite{bampomdp}. $\epsilon$-optimality  is established  for the
former case but not under model uncertainty.
%The Bayes-adaptive technique generates approximate observation and transition functions via online sampling.
%
Another    related    approach    is   Monte    Carlo    Q-Alternating
(MCQ-Alt)~\cite{mcqalt}, a quasi model-based RL method for Dec-POMDPs,
where agents  take turns learning given  the learned policy so  far of
the  other agent.  While no  model is  known a  priori, MCQ-Alt  first
estimates  model parameters  in the  intermediate step  and plans.  In
contrast, our methods  perform model-free RL in a  MPOMDP setting with
simultaneously learning agents.

Finally,     the    infinite     regional    policy     representation
(iRPR)~\cite{irpr}  performs model-free  exploration of  nonparametric
policies for POMDPs. While iRPR  allows an unbounded number of states,
its convergence is sample-intensive  requiring $10^3$ samples even for
the  simple   1D-maze  domain;   this  makes   it  a   poor  departure
point. Additionally, parameters must be manually configured to achieve
optima.  However,   it  does   outperform  the   model-based  infinite
POMDP~\cite{infpomdp}.

\vspace{-0.1in} 
\section{Heterogeneous Teams}
\label{sec:fr-posg}
\vspace{-0.1in} 

We focus on a system of  $Z$ heterogeneous agents cooperating toward a
common goal.  Specifically, $\{  \R_1, \R_2, \ldots,  \R_Z \}$  is the
collection of agent's reward functions, which may differ; this defines
the heterogeneity.   Here, $\R_i: S \times  A \rightarrow \mathbb{R}$,
$i \in  \mathcal{I}$ is an  agent's reward function, which  maps state
and joint action to value. An agent's reward $\R_i$ is decomposed into
local costs ($R_i$),  predicated on the joint physical  state and {\em
  individual} action,  and the global  reward ($R_G$), a  mapping from
the joint state and joint action, as shown below.
\vspace{-0.05in}
\begin{align}
	\R_i(s,a)=R_i(s,a_i)+R_G(s,a)
	\label{eqn:r}
\end{align}
Local costs may be unique to each agent and provide a way to model the
diversity  between   agents.  For   example,  the  global   reward  in
Fig.~\ref{fig:robot}  is 100  when the  robots are  aligned, otherwise
-1.  The  local  component  differs   between  robots  with  the  more
sophisticated platform  incurring a cost  of 2  for turning and  1 for
emitting infrared while it costs 3 for the other robot to do either.

%When a robot choose to stay at the current state(i.e. do nothing), it receives a reward 0. 

A trajectory  of $T$ steps  in this  {\em factored reward}  setting is
$\tau$=$(\vec{a}^0,$                             $\vec{\mathbf{r}}^0,$
$\vec{o}^1,  \vec{a}^1,   \vec{\mathbf{r}}^1,  \ldots,  \vec{o}^{T-1},
\vec{a}^{T-1},     \vec{\mathbf{r}}^{T-1})$.      Here,     $\vec{a}$,
$\vec{\mathbf{r}}$ and  $\vec{o}$ are vectors of  all agents' actions,
rewards  and   observations,  respectively.   An  agent's   reward  in
$\vec{\mathbf{r}}$, denoted by $\mathbf{r}_i$, is a {\em tuple} of the
local costs and global rewards received by $i$ based on the components
in Eq.~\ref{eqn:r}, and analogously for others.

\section{MCES for Factored Rewards}
\label{sec:mces}
\vspace{-0.1in}
We  introduce  two model-free RL approaches  for the factored-reward setting. Both generalize the \mcesp{} template.

\vspace{-0.05in}
\subsection{Joint Policy Iteration}
\label{subsec:mcesmp}
\vspace{-0.05in}
The class of  factored reward settings with agents  that cooperate may
be   cast  into   the  well-known   framework  of   multiagent  POMDPs
(MPOMDPs)~\cite{mpomdp}. These  generalize POMDPs to  multiple agents;
actions and  observations in a  MPOMDP are  a joint of  the individual
agent  actions and  observations. Importantly,  the output  is a  {\em
  single}  policy  that  maps  joint observation  sequences  to  joint
actions. %We define the MPOMDP below.
%\vspace{-0.1in}
\begin{align*} 
\text{MPOMDP} \triangleq \langle \mathcal{I}, S, A, \Omega, T, O, \R
  \rangle 
\end{align*}
\vspace{-0.1in}

%\begin{itemize}[topsep=0pt,leftmargin=*,itemsep=0pt]
%	\item $\mathcal{I}=\{1,...,Z\}$ is the set of $Z$ interacting agents;
%	\item $S$ is the set of physical states;
%	\item $A = A_1\times ... \times A_Z$,  is the set of joint actions where
%	$A_1$, $A_2$, \ldots, $A_Z$ are  the sets of each agent's actions. A
%	joint action is then, $\vec{a}=\{a_1,...,a_Z\}$;
%	\item  $\Omega =  \Omega_1 \times  ...\times\Omega_Z$, is  the  set of
%	joint observations where  $\Omega_1$, $\Omega_2$, \ldots, $\Omega_Z$
%	are the sets  of each agent's observations.  A  joint observation is
%	$\vec{o}=\{o_1,...,o_Z\}$;
%	\item $T  :S \times  A \times S  \rightarrow [0,1]$ is  the transition
%	function that  determines how the  state evolves. It maps  an origin
%	state, a joint action, and an arrival state to a probability;
%	\item  $O:  \Omega  \times  S  \times  A  \rightarrow  [0,1]$  is  the
%	observation   function  that  gives   the  informativeness   of  the
%	observations toward  the state. It maps an  observation, the arrival
%	state, and a joint action to a probability;
 %     \item  $\R:  S  \times  A  \rightarrow  \mathbb{R}$.  While  the
  %      factored reward setting also  includes the individual costs of
   %     each  agent's actions,  we obtain  the single  reward function
    %    that is needed as follows:
$\mathcal{I}=\{1,...,Z\}$ is the set of $Z$ interacting agents; $S$ is
the set of physical states; $A = A_1\times ... \times A_Z$, is the set
of joint  actions where $A_1$,  $A_2$, \ldots,  $A_Z$ are the  sets of
each     agent's    actions.     A    joint     action    is     then,
$\vec{a}=\{a_1,...,a_Z\}$;
$\Omega  = \Omega_1  \times ...\times\Omega_Z$,  is the  set of  joint
observations where $\Omega_1$, $\Omega_2$,  \ldots, $\Omega_Z$ are the
sets   of  each   agent's  observations.    A  joint   observation  is
$\vec{o}=\{o_1,...,o_Z\}$; $T :S \times  A \times S \rightarrow [0,1]$
is the transition  function that determines how the  state evolves. It
maps  an origin  state, a  joint  action, and  an arrival  state to  a
probability; $O:  \Omega \times S  \times A \rightarrow [0,1]$  is the
observation   function  that   gives   the   informativeness  of   the
observations toward  the state.  It maps  an observation,  the arrival
state,     and     a     joint    action     to     a     probability;
$\R: S  \times A \rightarrow  \mathbb{R}$.  While the  factored reward
setting also includes the individual costs of each agent's actions, we
obtain the single reward function that is needed as: \vspace{-0.05in}
\begin{align}
		\R(s,a)=\sum\nolimits_{i\in \mathcal{I}} R_i(s,a_i)+R_G(s,a)
		\label{eq:rz}
	\end{align}
\vspace{-0.1in}

%\vspace{-0.2in} 
%	Equation~\ref{eq:rz} characterizes the cumulative team reward.
%\end{itemize}
\setlength{\textfloatsep}{2pt}
\begin{algorithm}[t]
	\caption{\mcesposg{}}
	\label{alg:MCESMP}
	\begin{small}
	\begin{algorithmic}[1]
		\REQUIRE Q-value table initialized and initial joint policy, $\pi$, that is greedy w.r.t Q-values; learning rate schedule $\alpha$; error $\epsilon$; and horizon $T$
		\STATE Initialize count $c_{\vec{o}, \vec{a}} \gets 0$ for all $\vec{o}$ and $\vec{a}$,
		$m \gets 0$
		\REPEAT
		\STATE Pick joint observation history $\vec{o}$ and joint action $\vec{a}$
		\STATE Set $\pi$ to $\pi \gets (\vec{o}, \vec{a})$
		%    \FOR {$k$ iterations} \DO 
		\STATE Generate trajectory $\tau$ of length $T$ online by simulating transformed joint policy $\pi \gets (\vec{o}, \vec{a})$
		\STATE $Q_{\pi \gets (\vec{o}, \vec{a})} \gets (1-\alpha(m, c_{\vec{o}, \vec{a}})) \cdot Q_{\pi \gets (\vec{o}, \vec{a})} + \alpha(m, c_{\vec{o}, \vec{a}}) \cdot \R_{post-\vec{o}}(\tau)$
		\STATE $c_{\vec{o}, \vec{a}} \gets c_{\vec{o}, \vec{a}} + 1$
		%    \ENDFOR
		\IF {$\max_{\vec{a}'} Q_{\pi \gets (\vec{o}, \vec{a}')} > Q_{\pi} + \epsilon(m, c_{\vec{o}, \vec{a}}, c_{\vec{o},\pi(\vec{o})})$}
		\STATE $\pi(\vec{o}) \gets \vec{a}'$ where $\vec{a}' \gets \arg \max Q_{\pi \gets (\vec{o}, \vec{a}')}$   
		\STATE $m \gets m + 1$
		\STATE Reset $c_{\vec{o}, \vec{a}} \gets 0$ for all $\vec{o}$ and $\vec{a}$
		\ENDIF
		\UNTIL{termination}
	\end{algorithmic}
	\end{small}
\end{algorithm}

%Our  first   algorithm  straightforwardly generalizes  \mcesp{}, 
%in which we derive a canonical MPOMDP as described above.
Algorithm~\ref{alg:MCESMP},   which   we   call   MCES   for   MPOMDPs
(\mcesmp{}),  straightforwardly  generalizes  \mcesp{}.   It  modifies
\mcesp{} in  several ways. Line  3 picks joints instead  of individual
observations and  actions. In line  4, $\pi \gets  (\vec{o}, \vec{a})$
denotes  the {\em  transformed  policy} that  prescribes $\vec{a}$  on
encountering observation sequence $\vec{o}$.  The trajectory $\tau$ in
line 5  is as described  in the previous  section. Line 6  updates the
Q-value  by   $R_{post-\vec{o}}$  with  an  averaging   learning  rate
$\alpha(m,  c)   =  \frac{1}{c+1}$,  where   $m$  is  the   number  of
transformations  taken so  far and  $c$ the  count of  updates to  the
Q-function.  In this instance, $m$  does not affect the learning rate.
To define  $\R_{post-\vec{o}}(\tau)$ in line  6, we begin  by defining
$\R(\tau)$ as
%\begin{align}
$	\R(\tau)=\sum\nolimits_{t=0}^{T-1}  \gamma^t  \sum\nolimits_{i\in\mathcal{I}} r_i^t + r_G^t  $,
%	\label{eq:rval}
%\end{align}
where $r_i^t$ and  $r_G^t$ are the local and global  components of the
reward    $\mathbf{r}_i$    at    time   $t$    received    by    $i$.
$\R_{post-\vec{o}}(\tau)$  is  then  the portion  of  $\R(\tau)$  that
succeeds joint observation $\vec{o}$.

We may instantiate the \mcesmp{}  template using PALO bounds to obtain
an algorithm  \mcesmppac{} that  can be  implemented similarly  to the
PALO instantiation of \mcesp{}. A key difference from \mcesp{} is that
the policy maps joints observations to joint actions, due to which the
size of  the local policy  neighborhood $N^{\sf MP}$  is significantly
larger. Specifically, \vspace{-0.05in}
\begin{small}
	\begin{align}
		N^{\textsf{MP}}=\prod_{i\in\mathcal{I}}|A_i|
          \left(\frac{\prod_{i\in\mathcal{I}}|\Omega_i|^T-1}{\prod_{i\in\mathcal{I}}|\Omega_i|-1}-1\right).
        \label{eq:N-MP}
	\end{align}
\end{small}
The other  difference is in  the maximal  range of action-values  of a
policy   $\pi$    and   its   transformation   $\pi'$,    denoted   as
$\Lambda(\pi,\pi')$ previously. We now utilize the maximum and minimum
values  of  the reward  function  defined  in Eq.~\ref{eq:rz}  in  the
computation of $\Lambda$ in Eq.~\ref{eq:lambda}.  Given these changes,
the definitions of $k_m$ and $\epsilon(m,p,q)$ as in Eqs.~\ref{eq:k_m}
and~\ref{eq:epsilon}  modify to  accommodate them,  and the  algorithm
terminates similarly.  \mcesmppac{} may  terminate early  similarly to
\mcesp{}.
\begin{prop}[Proof follows from Greiner~\cite{PALO}]
	\label{prop:MCESMP}
	With  probability  $1-\delta$,  \mcesmppac{} iterates  over  a
        series  of  policies  mapping   joint  observations  to  joint
        actions, $\pi^1,\pi^2,\ldots,\pi^m$, such that the transformed
        policy $\pi^{j+1}$  dominates the  previous policy  $\pi^j$ in
        value    for    all    agents    and    terminates    to    an
        \textit{$\epsilon$-locally optimal}  policy $\pi^m$,  where no
        neighbor  dominates   the  converged   policy  by   more  than
        $\epsilon$.
\end{prop}
%\noindent   Proof   of   this   proposition  follows   directly   from
%Greiner~\cite{PALO}.

\vspace{-0.05in} 
\subsection{Joint Transformation of Individual Policies}
\label{subsec:mcesfp}
\vspace{-0.05in}  Motivated  by   previous  approaches  in  multiagent
planning  that divide  the joint-policy  search space  into individual
agent     policy     search     spaces     with     a     coordination
mechanism~\cite{Amato07:Optimizing,factoredmdp},  our   second  method
seeks to  learn a vector  of policies, one  for each agent.   A policy
$\pi_i$ for an agent $i$ in this vector maps joint observations of all
agents to the action prescribed for $i$. It does not require combining
the rewards as in Eq.~\ref{eq:rz}; rather it continues to utilize each
agent's   separate   reward   signal  $\mathbf{r}_i$   obtained   from
$\mathcal{R}_i$.     The     model-free    RL    is     outlined    in
Algorithm~\ref{alg:MCESFP},   and  we   refer  to   it  as   MCES  for
factored-reward MPOMDPs (\mcesfp{}).

Similar to \mcesmp{}, we begin  by picking a joint observation history
$\vec{o}$  and action  $\vec{a}$  either randomly  or  in an  iterated
manner.   However, it  uses these  to transform  each agent's  current
policy by setting the action at $\vec{o}$ with its action in $\vec{a}$
(line 4). \mcesfp{} maintains the Q-value for each agent's transformed
policy  (line 7)  {\em  additionally  indexed by  the  joint of  other
  agents'  actions picked  for  $\vec{o}$}.   This ensures  consistent
updates of  Q-values for the same  set of joint actions,  and thus the
multiagent policy vector. As such, it maintains as many Q-functions as
the number of agents and  combinations of other agents' actions picked
for $\vec{o}$  in the  worst case, i.e.,  $\mathcal{O}(ZA^{Z-1})$.  To
obtain    $\R_{i,post-\vec{o}}(\tau)$   in    line   7    note   that,
$\mathcal{R}_i(\tau)=\sum\nolimits_{t=0}^{T-1}\gamma^t(r_i^{t}+r_G^{t})$
Then,   $\R_{i,post-\vec{o}}(\tau)$   is   simply   the   portion   of
$\mathcal{R}_i(\tau)$ that obtains after $\vec{o}$ has occurred in the
trajectory.  Finally, the  conjunction  on line  10  ensures that  all
transformations are accepted together or none are.

	\begin{algorithm}[!ht]
		\caption{\mcesfp{}}
		\label{alg:MCESFP}
		\begin{small}
		\begin{algorithmic}[1]
			\REQUIRE Q-value  tables initialized and initial  profile of agent
			policies, $\{\pi_i\}_{i=1}^Z$,  that are greedy  w.r.t.  Q-values;
			learning rate schedule $\alpha$; error $\epsilon$; and horizon $T$
			
			%    \FORALL {$i\in\mathcal{I}$}
			%    \FORALL {$\vec{o}_i,a_i$}
			\STATE Initialize $c^i_{\vec{o}, a_i} \gets 0$ for all $\vec{o}$, $a_i$, and
			$i \in \mathcal{I}$, 
			%    \ENDFOR
			%    \ENDFOR
			$m  \gets 0$ \REPEAT \STATE Pick  joint observation history
			$\vec{o}$ and  joint action $\vec{a}$  
			%\FORALL {$i\in\mathcal{I}$}
			\STATE Set $\pi_i$ to neighboring policy $\pi_i \gets (\vec{o}, a_i)$ for all {$i\in\mathcal{I}$}
			%\ENDFOR
			%    \STATE Generate trajectory $\tau$ (=$\{\tau_i\}_{i=1}^Z$) of
			%    length $T$ by simulating all transformed policies $\{\pi_i \gets
			%    (\vec{o}, a_i)\}_{i=1}^{Z}$
			\STATE Generate trajectory $\tau$ of length $T$ online by
			obtaining each agent's action using its transformed policy 
			$\pi_i \gets (\vec{o}, a_i)$ for each $i \in \mathcal{I}$
			\FORALL {$i\in\mathcal{I}$}
			\STATE $Q^{\vec{a}_{-i}}_{\pi_i \gets (\vec{o}, a_i)} \gets (1-\alpha(m, c^i_{\vec{o}, a_i})) \cdot Q^{a_{-i}}_{\pi_i \gets (\vec{o}, a_i)} +\alpha(m, c^i_{\vec{o}, a_i}) \cdot \R_{i,post-\vec{o}}(\tau)$
			\STATE $c^i_{\vec{o}, a_i} \gets c^i_{\vec{o}, a_i} + 1$
			\ENDFOR
			\small \IF {{ $\bigwedge\limits_{i \in \mathcal{I}} \left
					(\max\limits_{a_i' \in \mathcal{A}_i}   Q^{\vec{a}_{-i}}_{\pi_i
						\gets (\vec{o}, a'_i)} > Q^{\vec{a}_{-i}}_{\pi_i}+\epsilon(n,
					c^i_{\vec{o},a_i}, c^i_{\vec{o}, \pi_i(\vec{o})})\right )$}}
			\normalsize
			% \FORALL {$i\in\mathcal{I}$}
			\STATE $\pi_i(\vec{o}) \gets a'_i$ where $a'_i \gets \arg \max
			Q^{a_{-i}}_{\pi \gets (\vec{o}, a'_i)}$ $\forall$$i \in \mathcal{I}$   
			\STATE $m \gets m + 1$
			\STATE Reset $c^i_{\vec{o}, a_i} \gets 0$ for all $\vec{o}$, $a_i$, and
			$i \in \mathcal{I}$
			\ENDIF
			\UNTIL{termination}
		\end{algorithmic}
		\end{small}
	\end{algorithm}

        {\em  Consequently, \mcesmp{}  could approach  the same  local
          optima in policies as \mcesfp{},  but may follow a different
          path.}  Specifically,   if  any   agent  receives   a  worse
        individual cost,  despite the cumulative reward  being better,
        \mcesfp{} will  not transform. While this  may preclude higher
        team rewards in  interim steps, the benefit  is that \mcesfp{}
        targets  policies with  higher joint  values and  takes larger
        steps    in    its    search     as    we    demonstrate    in
        Section~\ref{sec:experiments}.

 When  instantiated  with  PALO   bounds  (\mcesfppac{}),  the  policy
 decomposition   approach  provides   significant  sample   complexity
 reductions.   This  is primarily  because  of  a much  reduced  local
 neighborhood,  where  the  first  factor  involving  actions  is  not
 exponential in the number of agents (c.f. Eq.~\ref{eq:N-MP}):
\begin{small}
	\begin{align*}
		N^{\textsf{FMP}}=|A_i|\left(\frac{\prod_{i\in\mathcal{I}}|\Omega_i|^T-1}{\prod_{i\in\mathcal{I}}|\Omega_i|-1}-1\right).
	\end{align*}
\end{small}
The conjunction on line 9 of \mcesfppac{} redefines
Eqs.~\ref{eq:k_m} and~\ref{eq:epsilon} as,
\begin{small}
	\begin{align}
		k_m = \ceil*{2\left(\frac{\Lambda(\pi_i,\pi'_i)}{\epsilon}\right)^2
			ln\left(\frac{\sqrt[2Z]{4Z-2}N^{\sf FMP}}{\sqrt[2Z]{\delta_m}}\right)}
		\label{eq:k_m_fp}
	\end{align}
	\begin{align}
          \epsilon^*(m,p) =
          \frac{\Lambda(\pi_i,\pi'_i)}{\sqrt{2p}}\sqrt{ln\frac{\sqrt[2Z]{(4Z-2)(k_m-1)}N^{\sf
          FMP}}{\sqrt[2Z]{\delta_m}}}.
		\label{eq:epsilon_fp}
	\end{align}
\end{small} 
Here, the  maximum range  of action values  $\Lambda(\pi_i,\pi'_i)$ is
computed analogously  to Eq.~\ref{eq:lambda}  with the change  that we
utilize  the maximum  and minimum  values  of the  reward function  in
Eq.~\ref{eqn:r}. This range  could get much narrower  in comparison to
the maximum range for \mcesmppac{}  because the rewards for the latter
are a  sum of  all local  costs. Of course,  this benefits  the sample
bound $k_m$
 %The following proposition states this formally:
\begin{prop}
  The sample  bound for  \mcesfppac{} given in  Eq.~\ref{eq:k_m_fp} is
  less than the  sample bound for \mcesmppac{} if  the following holds
  for all agents:
	\label{prop:kmfpmp}
	\begin{small}
		\begin{align*}
			ln\left(\frac{\sqrt[2Z]{4Z-2}~N^{\sf FMP}}{\sqrt[2Z]{\delta_m}}\right)
			<
			\left(1+\frac{\sum_{i\in\bar{\mathcal{I}}}\R_{i,max}}{\R_{i,max}+\R_{G,max}}\right)^2
			ln\frac{2N^{\sf MP}}{\delta_m}
		%\label{eq:k_m_diff}
		\end{align*}
              \end{small}
	where $\bar{\mathcal{I}}$ is the set  of all agents other than
        agent $i$.
\end{prop}
Proposition~\ref{prop:kmfpmp} is derived  by simplifying the condition
when  the  sample  bound  for  \mcesfppac{}  (Eq.~\ref{eq:k_m_fp})  is
smaller   than    that   of   \mcesmppac{}    (Eq.~\ref{eq:k_m}   with
$\Lambda=\Lambda(\pi,\pi')$   and   $N=N^{MP}$).    \mcesfppac{}   may
terminate early if no neighboring policy exceeds the current value by
\noindent $\epsilon-\epsilon^*(m,p)$.  %, $k_m=\ceil*{2\left(\frac{\Lambda(\pi,\pi')}{\epsilon}\right)^2 ln\frac{2N^{MP}}{\delta_m}}$.
Its behavior is characterized by the following proposition.
\begin{prop}
	\label{prop:MCESFP}
	With probability  $1-\delta$, \mcesfp{} iterates over  a joint
        set  of  individual  policies mapping  joint  observations  to
        individual  actions   $\pi^1,\pi^2,\ldots,\pi^m$  where  every
        individual          policy         in          the         set
        $\pi^{j+1}=\langle  \pi_1,\pi_2,\ldots,\pi_Z  \rangle$ in  the
        neighborhood of $\pi^j$ dominates each agent's previous policy
        in value  and terminates to an  $\epsilon$-locally optimal set
        of policies $\pi^m$. Here no  neighbor for any agent dominates
        the converged  policy by more than  $\epsilon$ without another
        agent receiving a worse reward.
\end{prop}

In   proving  Proposition~\ref{prop:MCESFP},   the  space   of  errors
increases multiplicatively  with the  number of agents.   In \mcesp{},
the agent may make two  categories of errors: \textit{transforming} or
\textit{terminating}  erroneously due  to noisy  samples of  the local
neighborhood.  In \mcesfp{}, one or  more agents may individually make
these errors, growing the number of errors by $2Z$.  The neighborhoods
are factored into $Z$ individual neighborhoods. These observations and
the values  in Eqs.~\ref{eq:k_m_fp} and~\ref{eq:epsilon_fp}  result in
the       bound       $\delta$.       Exhaustive       proofs       of
Propositions~\ref{prop:kmfpmp}  and~\ref{prop:MCESFP} are  included in
the supplementary material.
%an online appendix~\cite{supp_3695}.
%\begin{figure}[!ht]
%	\centerline{
%		\includegraphics[width=2.75in]{Figs/mpfpfig.png}}
%	\caption{\small Illustration of our methods in the factored-reward setting.}
%	\label{fig:mpfpfig}
%\end{figure}

%In summary, the two approaches for  the factored-reward setting are illustrated in %Fig.~\ref{fig:mpfpfig}. 
In summary,  \mcesmp{} explores policies which  map joint observations
to joint actions, aggregating all  local costs with the global reward.
\mcesfp{}  explores  the set  of  independent  policies mapping  joint
observations  to individual  actions, where  agents receive  their own
local cost  with the global  reward. The latter yields  reduced sample
complexity when instantiated with PALO guarantees.

\vspace{-0.05in} 
\section{Joint Policy Search Space Pruning}  
\vspace{-0.1in} 
PAC RL requires that we have  $k_m$ samples of each transformations in
the local neighborhood  to terminate, even those  of observations that
are less probable; such sequences may  be due to noisy observations by
one or more  agents.  Naturally, we may seek to  avoid evaluating such
transformations in  order to speed  up the overall  search.  Foregoing
these   portions   of   the  local   neighborhood   risks   preventing
\textsf{MCES}  from  transforming  to  a higher  value  joint  policy,
resulting in a loss of  expected value referred to as \textit{regret}.
Regret depends  on two factors: the  range of possible rewards  on the
pruned observation sequence, and its likelihood of occurring.  Even if
the range of rewards is equivalent  to other sequences, the regret due
to avoiding the sequence reduces as it grows more improbable.

We may  integrate this  pruning into  both Algorithms~\ref{alg:MCESMP}
and ~\ref{alg:MCESFP}  as suggested  by Ceren et  al.~\cite{mcesip} by
taking an additional input parameter, $\phi$,  that is a user bound on
allowable regret,  and maintaining a slowly  growing set $\mathcal{P}$
of  joint   observation  sequences  that   will  be  avoided.    If  a
transformation is sought on  an observation sequence in $\mathcal{P}$,
then it is skipped.  An observation sequence $\vec{o}$ is added to the
set if the cumulative regret  due to all sequences including $\vec{o}$
in $\mathcal{P}$ is still less than $\phi$.

We ensure that sufficient samples containing $\vec{o}$, $c_{\vec{o}}$,
are   obtained   before  its   regret   is   considered  by   defining
$\rho(c_{\vec{o}})$  as  $0$  if $c_{\vec{o}}\geq  \frac{k_m}{2}$  and
$+\infty$ otherwise. Then, $\rho$ is simply added to $\phi$.
%The algorithm for \mcesmp{} is modified in a similar way.

\vspace{-0.1in} 
\section{Experiments}
\label{sec:experiments}
\vspace{-0.1in}   

We  evaluate  \mcesmppac{}  and  \mcesfppac{} on  three  domains  with
heterogeneous agents  (see Table~\ref{tab:Domain}).  Our  first domain
is the well-known team  Tiger problem with 2 agents~\cite{tigerpomdp}.
We expand this  domain to include factored rewards:  the global reward
corresponds  to the  original reward  function, and  additionally each
agent $i$ incurs a cost of $i$  for opening a door.  The second domain
is  the  3-agent  Firefighting previously  introduced  for  evaluating
cooperative decentralized planning~\cite{amato}  using $4$ houses with
a maximum  fire intensity of $3$.   Agents incur two local  costs: the
distance between  houses if  they move  ($\text{distance}/(10-i)$) and
the fire intensity of the  target ($f_h/(10-i)$).  Our final domain is
the 2- and 4-agent robot alignment problem~\cite{alignment} where each
agent has two states and four states respectively.  Robots incur local
costs on turning ($-2-i$) and emitting IR ($-1-i$).
%Table~\ref{tab:Domain}     specifies     the    domain     parameters.

\begin{table}[!ht]
  \vspace{-0.1in}
	\centering
	\begin{small}
	\begin{tabular}{| c | c |}
          \hline
          \textbf{Domain} & \textbf{Specifications
                            (for all $\epsilon=0.1$, $\delta=0.1$)} \\
          \hline
          2-agent Team Tiger ({\sf Tiger}) & $|S|=2$, $|A_i| =
                                             3$, $|\Omega_i| =
                                             2$, $T=5,6$,
                                             \textsf{Opt=25.64, 32.11}\\
		% \hline
		% Team Tiger T=4 & $\epsilon=0.125$, $\delta=0.075$, $T=4$, $Opt=18.11$\\
          \hline
          3-agent Firefighting ({\sf Fire}) & $|S|=755$, $|A_i|=3$, $|\Omega_i|=2$, $T=3,4$,
                                              \textsf{Opt=-5.94, -4.83}\\
          \hline
          2- and 4-robot alignment ({\sf Align}) & $|S|=4$,
                                                   $|A_i|=4$, $|\Omega_i|=2$, $T=4$,
                                                   \textsf{Opt=14.39, 24.72}\\
          \hline

%		4-agent robot alignment & $\epsilon=0.175$, $\delta=0.1$, $T=4$,,
%		\textsf{Opt=24.72}\\
%		\hline
	\end{tabular}%
	\end{small}
	\caption{\small  Parameters  for   the  problem  domains  with
          optimal policy  values.  Our experiments involve  domains of
          various  sizes,   and  increasing  numbers  of   agents  and
          horizons.}
	\label{tab:Domain}%
        \vspace{-0.15in}
\end{table}

In each domain, agents simultaneously  act in a sequential environment
with private observations  that are conveyed exactly  and perfectly to
the centralized  learner. Agents  perform their  actions based  on the
joint  observation  sequence  as  prescribed by  the  policy.   Entire
trajectory  of $T$  time  steps  is then  sent  back  to the  learning
algorithm. Each  trial is initialized  with a random pairing  of joint
observation sequences to  joint actions in \mcesmp{}  or to individual
actions in \mcesfp{}.

\begin{table*}[!ht]
	\begin{small}
	\centering
	\scalebox{1}{
		\begin{tabular}{|c|c|c|c|c|c|c|c|c|}
                  \cline{4-9}
                  \multicolumn{3}{c|}{ } & \textbf{Init. Value} & \textbf{Final Value} & \textbf{Samples} & \textbf{Transforms} & \textbf{$k_m$} & \textbf{$\phi$} \\
                  \cline{4-9}
                  \hline
                  \multirow{6}{*}{\begin{turn}{90}\textbf{\textsf{FMP+PAC} }\end{turn}} &\textbf{2-agent} & \textbf{T=5} & -207.2 $\pm$ 1.2 & 20.8 $\pm$ 0.9 & 32,594 $\pm$ 227 & 37 $\pm$ 1.1 & 79,482 & 0.1 \\
      \cline{3-9}
      &\textbf{Tiger}& \textbf{T=6} & -279.8 $\pm$ 3.1 & 26.4 $\pm$ 0.4 & 67,914 $\pm$ 104 & 59 $\pm$ 1.2 & 144,582 & 0.1 \\
      \cline{2-9}
      &\multicolumn{2}{|l|}{\textbf{2-robot Align}} & -210.7 $\pm$ 0.7 & 11.4 $\pm$ 0.6 & 28,139 $\pm$ 31 & 16 $\pm$ 1.4 & 56,838 & 0.1 \\
      \cline{2-9}
      &\textbf{3-agent} & \textbf{T=3} & -10.5 $\pm$ 0.1 & -6.4 $\pm$ 0.3 & 20,921 $\pm$ 93 & 3.4 $\pm$ 0.5 & 40,580 & 0.1 \\
      \cline{3-9}
      &\textbf{Fire}& \textbf{T=4} & -12.8 $\pm$ 0.1 & -5.8 $\pm$ 0.2 & 46,235 $\pm$ 95 & 40 $\pm$ 1.8 & 108,512 & 0.1 \\
      \cline{2-9}     
      &\multicolumn{2}{|l|}{\textbf{4-robot Align}} & -246.0 $\pm$ 1.2 & 20.1 $\pm$ 0.2 & 39,256 $\pm$ 180 & 36.8 $\pm$ 0.4 & 84,651 & 0.1 \\               
      \hline

      \hline
      \multirow{6}{*}{\begin{turn}{90}\textbf{\textsf{MP+PAC\phantom{.}}}\end{turn}}
                                         &\textbf{2-agent} & \textbf{T=5} & -207.2 $\pm$ 1.2 & 6.7 $\pm$ 1.1 & 33,985 $\pm$ 183 & 22 $\pm$ 1.4 & 80,484 & 0.2 \\
                  \cline {3-9}
                                         &\textbf{Tiger} & \textbf{T=6} & -279.8 $\pm$ 3.1 & 8.6 $\pm$ 0.5 & 69,904 $\pm$ 177 & 36 $\pm$ 1.7 & 143,883 & 0.2 \\
                  \cline{2-9}
                                         &\multicolumn{2}{|l|}{\textbf{2-robot Align}} & -210.7 $\pm$ 0.7 & 2.7 $\pm$ 0.3 & 29,124 $\pm$ 32 & 10 $\pm$ 1.2 & 59,242 & 0.2 \\
                  \cline{2-9}
                                         &\textbf{3-agent} & \textbf{T=3} & -10.5 $\pm$ 0.1 & -8.6 $\pm$ 1.2 & 21,940 $\pm$ 122 & 2.2 $\pm$ 0.4 & 53,904 & 0.2 \\
                  \cline {3-9}
                                         &\textbf{Fire} & \textbf{T=4} & -12.8 $\pm$ 0.1 & -9.3 $\pm$ 0.1 & 48,216 $\pm$ 155 & 26 $\pm$ 1.2 & 121,507 & 0.2 \\
                  \cline{2-9}
                                         &\multicolumn{2}{|l|}{\textbf{4-robot Align}} & -246.0 $\pm$ 1.2 & 6.9 $\pm$ 1.3 & 39,940 $\pm$ 101 & 33 $\pm$ 0.5 & 111,829 & 0.2 \\
      \hline
				
			\hline
			\multirow{6}{*}{\begin{turn}{90}\textbf{\textsf{BA
                                FV-POMCP} }\end{turn}} &\textbf{2-agent} & \textbf{T=5} & -207.2 $\pm$ 1.2 & -55 $\pm$ 1.7 & 32,993 $\pm$ 95 & - & - & - \\
			\cline{3-9}
			&\textbf{Tiger}& \textbf{T=6} & -279.8 $\pm$ 3.1 & 5.8 $\pm$ 0.2 & 67,884 $\pm$ 63 & - & - & - \\
			\cline{2-9}
			&\multicolumn{2}{|l|}{\textbf{2-robot Align}} & -210.7 $\pm$ 0.7 & -20.9 $\pm$ 0.4 & 26,169 $\pm$ 18 & - & - & -\\
			\cline{2-9}
			&\textbf{3-agent} & \textbf{T=3} & -10.5 $\pm$ 0.1 & -8.7 $\pm$ 0.2 & 20,962 $\pm$ 80 & - & - & - \\
			\cline{3-9}
			&\textbf{Fire}& \textbf{T=4} & -12.8 $\pm$ 0.1 & -9.2 $\pm$ 0.2 & 46,270 $\pm$ 84 & - & - & - \\
			\cline{2-9}			
			&\multicolumn{2}{|l|}{\textbf{4-robot Align}}
      &-246.0 $\pm$ 1.2 & 7.7 $\pm$ 0.1 & 40,040 $\pm$ 93 & - & - & - \\
      \hline

		\end{tabular}}
		\end{small}
		\vspace{-0.05in}
                \caption{ \small Average metrics  with std. error over
                  5 runs for all methods.  In every case, \mcesfppac{}
                  significantly   outperforms   \mcesmppac{}   and   a
                  benchmark in value of  converged policy while taking
                  similar (or fewer) samples per transform.  We report
                  the  team  reward  of   the  initial  and  converged
                  policies for all methods.  }
	\label{tbl:avgs}\vspace{-0.1in}
\end{table*}

\vspace{-0.1in}
\noindent    \paragraph{Comparative    performances}    Factored-value
POMCP~\cite{amato} offers  the previous  best model-based  solution of
MPOMDPs.   It may  be combined  with the  Bayes adaptive  approach for
initial  model  building.   We empirically  validate  the  drastically
reduced  number  of  samples  required  by  \mcesfppac{}  compared  to
\mcesmppac{}.   We  sought  to   obtain  high-valued  {\em  converged}
policies from  the two methods  for the  least regret in  a reasonable
amount of  clock time (each  run was capped at  10 hours on  a typical
PC).  Additionally, the sample counts  of the two methods are compared
with  those used  by BA-FV-POMCP.   The latter  uses samples  from the
environment for  the initial model-building phase  and simulations for
the model solving.   As the simulations are performed  using the built
model, we  limit our attention to  the number of samples  utilized for
model  building.  We  report  on  the values  of  the converged  joint
policies  for  all  three  methods  while  allowing  \mcesmppac{}  and
BA-FV-POMCP to  use a similar  number of samples as  \mcesfppac{}.  To
permit comparison, values of policies  from both methods were computed
similarly: the policies were simulated and local rewards of all agents
were  summed and  added to  the  global reward;  this was  accumulated
across  all  steps  in  the   trajectory.   As  another  baseline,  we
implemented  an  algorithm that  uses  the  single-agent \mcesp{}  for
learning the policy  of each agent individually, but  agents are still
situated  in the  multiagent setting.   Trajectories are  generated by
jointly performing  the actions  from each agent's  policy. Experiment
data is included in the supplementary material.

\begin{figure}[!ht]\vspace{-0.05in}
		\begin{minipage}{2.75in}
			\centerline{\includegraphics[width=2.5in]{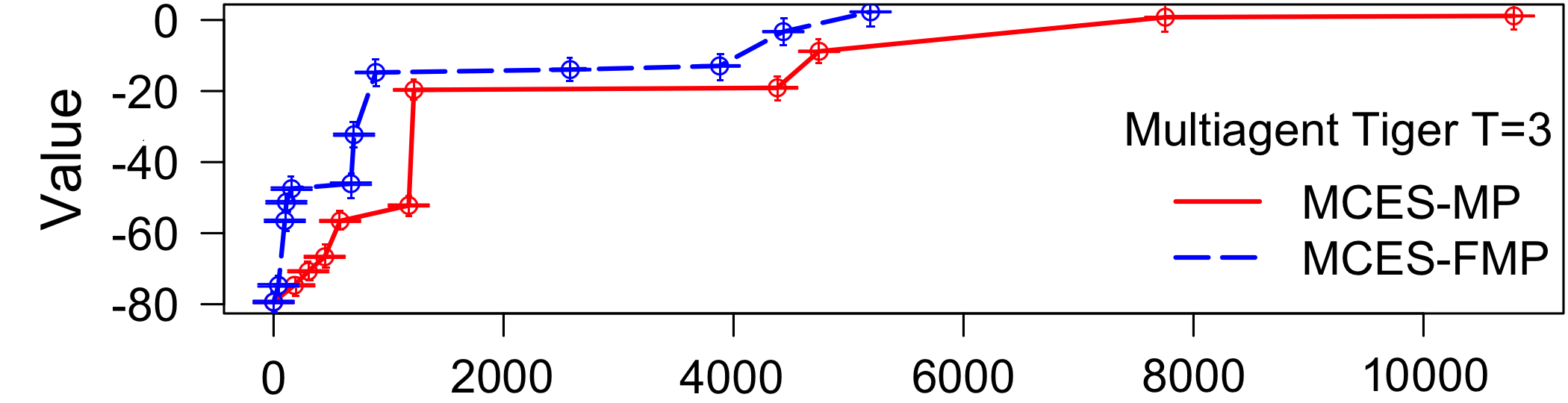}}
			\centerline{\includegraphics[width=2.5in]{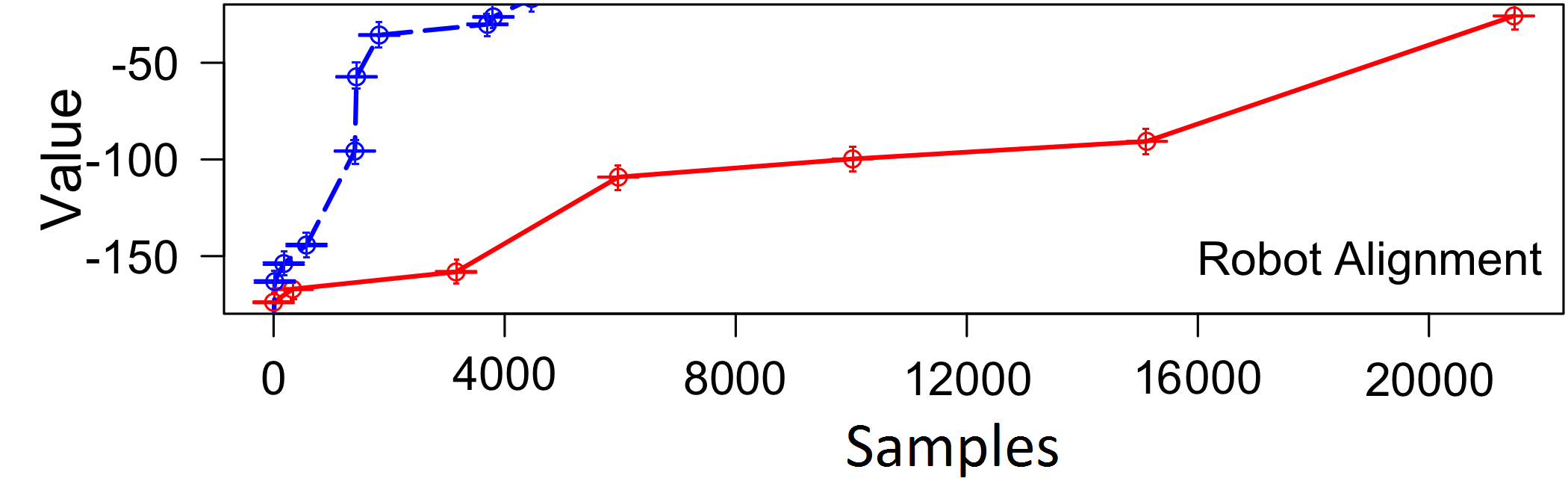}}
		\end{minipage}
		\begin{minipage}{2.75in}
			\centerline{\includegraphics[width=2.5in]{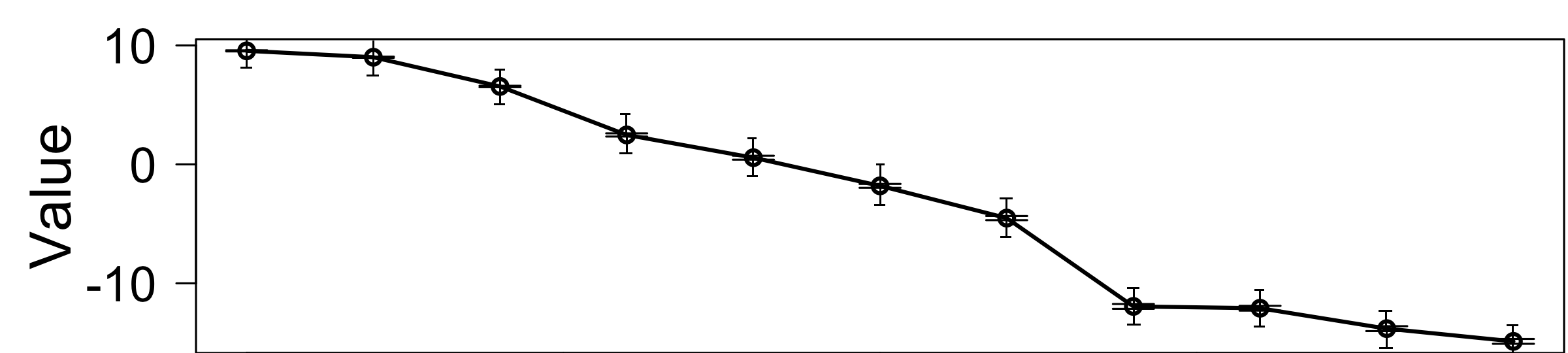}}
			\centerline{\includegraphics[width=2.5in]{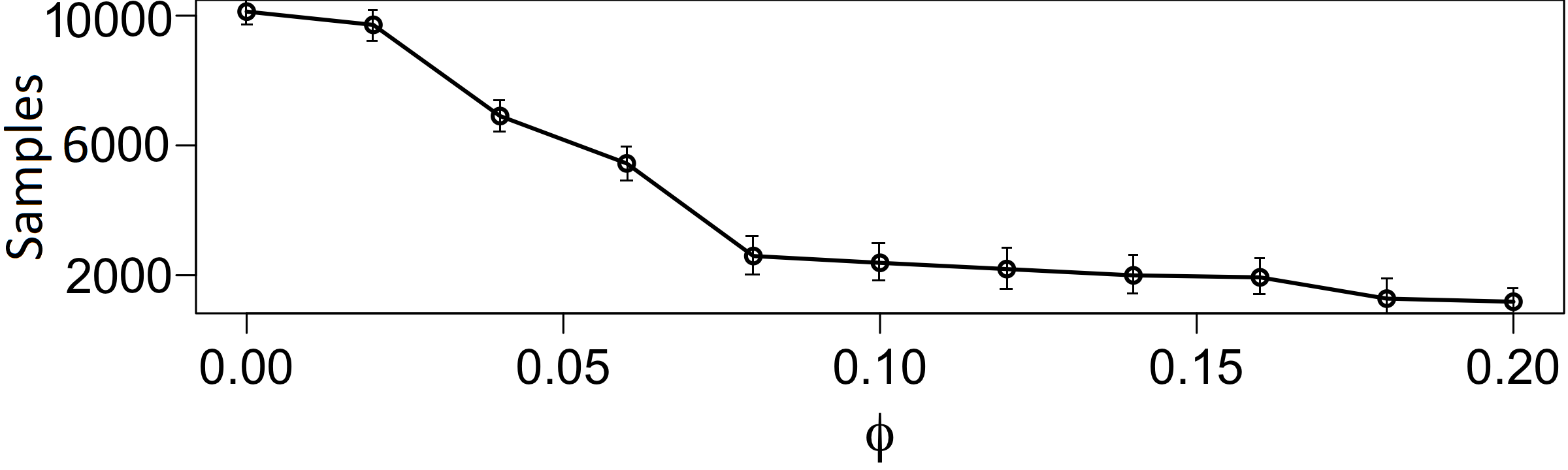}}
		\end{minipage} \vspace{-0.05in}
                \caption{\small$({\bf   left})$  Intermediate   policy
                  values  for  team   Tiger  $T=3$  ($\phi=0.05$)  and
                  2-agent alignment  ($\phi=0.1$). \mcesfp{} converges
                  to similar  or better policies than  \mcesmp{} using
                  far fewer samples.   $({\bf right})$ Illustration of
                  \mcesfp{} with varying $\phi$, including convergence
                  to near-optimal  values with no pruning.   Each data
                  point averages 11 runs  on team Tiger T=3.  Standard
                  errors are very small.}
            \vspace{-0.07in}
	\label{fig:transform}
\end{figure}

% The  algorithms' parameters  are set  in order  to make  the algorithm
% converge within the time  limit. We set a limit of  60 minutes for the
% team Tiger $T=3$;  300 when $T=4$; 300 for the  2-robot alignment; 450
% minutes for  the 3-agent Firefighting  $T=3$; 800 when $T=4$;  and 600
% for  the  4-robot alignment.  These  limits  are sufficient  to  allow
% learning high-valued policies.  Within this time,
Table~\ref{tbl:avgs} lists the metrics with the values averaged over 5
runs of  each algorithm  starting at  differing initial  policies. The
$\phi$  that  led  to  convergence is  also  reported.   Observe  that
\mcesfppac{} yields a  joint policy that is  significantly better than
the joint from \mcesmppac{} and BA-FV-POMCP for about the same numbers
of  samples.   This  holds  for multiple  problem  domains  and  their
configurations. Furthermore, the policies  generate coordination as is
evident  from   the  improvement  over  the   \mcesp{}  baseline.   In
Fig.~\ref{fig:transform} (left), we show  progressions of the Q-values
of  the  joint  policies  as  they  transform  in  both  methods.   In
particular,  transforms by  \mcesfppac{}  are more  rewarding and  the
difference in samples is dramatic for robot alignment.

\vspace{-0.1in}
\noindent  \paragraph{Characterization of  local optima}  \mcesfppac{}
converges  to good  quality policies  that are  close to  optimal.  To
illustrate this,  we let the allowable  regret $\phi$ vary from  0 (no
pruning)  to   0.2  in  the   context  of  {\sf  Tiger}   with  $T=3$.
Figure~\ref{fig:transform} (right)  shows a  converged value  of 10.28
for the policy vector in the  absence of pruning; this gradually drops
as $\phi$  increases.  Expectedly,  the number  of used  samples drops
significantly as well.  With  $\phi=0$, \mcesfppac{} achieved policies
of value  $-6.26$ for {\sf Fire}  and $11.81$ for {\sf  Align}, albeit
with more  samples.  Figure~\ref{fig:robot} demonstrated  a successful
run on the particularly challenging 2-robot {\sf Align} as evidence of
the  good-quality policies  learned by  \mcesfp{}. The  learned policy
vector guides the two robots to a successful alignment.

% While \mcesmp{} may show a  very minor gain over \mcesfp{}'s converged
% value,  it  does  so  with  a  significantly  higher  required  sample
% count. In every case, \mcesfp{} takes half to a third of the time with
% \textit{at most}  a loss  of $2.25\%$ of  the final  reward. \mcesfp{}
% largely identifies  similar transformations  to \mcesmp{},  but misses
% those that  require a loss in  reward for one  agent in lieu of  a far
% greater team reward.

%However, despite this shortcoming, \mcesfp{} illustrates an efficient approach to arriving at good local optima for factored-reward MPOMDPs.
\vspace{-0.1in}
\section{Concluding Remarks}
\label{sec:conclusion}
\vspace{-0.1in}  

\mcesp{} offers  elegant policy-based RL in  the partially observable,
single-agent context.  We generalized \mcesp{} to a heterogeneous team
setting, introducing model-free learning by  searching in the space of
joint policies.  We presented two templates for which
%\mcesmp{},  which casts  the problem  as a
%canonical  MPOMDP mapping  joint  observations to  joint actions,  and
%\mcesfp{},  which decomposes  the joint  into individual  policies for
%each  agent,  which  map  joint  observations  to  individual  agent's
%actions. %The presented methods are efficient than previous works. 
PAC-like guarantees  were established on  the local optimality  of the
converged policies.  Empirical results on three domains with number of
agents  ranging from  2 to  4 comprehensively  establish the  positive
performance  of  these  first   model-free  techniques  that  fill  an
important gap in the literature on cooperative decision
making. % Of course, model-based learning for MPOMDPs~\cite{amato} can
% show improved scalability. While our primary objective, in this paper,
% was  to establish  and  reduce  sample complexity,  we  will focus  on
% further scaling the number of agents in subsequent work.

Some  interesting  observations can  be  made  about the  relationship
between  optima  in  \textsf{-FMP}  and \textsf{-MP}.  All  optima  in
\mcesmp{}  are also  optima in  \mcesfp{}, as  a reduced  total reward
reflects   a  decrease   in  individual   reward  for   some  or   all
agents. However, a neighboring policy's  total reward may be higher if
the cost  for one  agent reduces  more than  that incurred  by another
agent.  In   this  case,   \textsf{-FMP}  will  not   transform  while
\textsf{-MP} will. Therefore,  optima in {\sf -FMP} may not  be so for
{\sf  -MP}. Even  so, in  practice,  \textsf{-FMP} may  route to  this
policy   via  a   different   path  or   arrive  at   similarly-valued
optima. While both methods search through the same joint policy space,
they differ in convergence criteria.

While our method's scalability is  fundamentally limited by the growth
of $N^{\text{FMP}}$ and the sample  guarantees imposed by PALO bounds,
it is  noteworthy that when  the amount  of interaction with  the real
environment is roughly equal,  \mcesfp{} significantly outperforms the
state-of-the-art   method,  BA   FV-POMCP,  while   being  model-free.
Consequently, the  relative scalability  of BA  FV-POMCP comes  at the
expense  of  any sample  guarantees  as  well as  actual  performance.
Importantly,  BA  FV-POMCP's  scalability  may  not  confer  pragmatic
benefits, because real-world interactions often dominate learning time
(consider  a  mobile robot  collecting  samples  using typically  slow
actuators).  When  judged from  this holistic perspective,  this paper
significantly advances the frontier of  multiagent systems.  As a next
step,  we  are   studying  recent  advances  in  using   deep  RL  for
POMDPs~\cite{Hausknecht15:Deep},  and the  challenges in  generalizing
the network and samples to the exponentially-harder MPOMDPs.

%As \textsf{-FMP} rejects a transformation where any individual agent receives a lower reward, and \textsf{-MP} transforms based on total reward, not all optima in one method are shared by the other.

% \input{appendix}

%%%%%%%%%%%%%%%%%%%%%%%%%%%%%%%%%%%%%%%%%%%%%%%%%%%%%%%%%%%%%%%%%%%%%%%%%%%%%%%%%%%%%%%%%%%%%%%%%%%%%%%%%
%% bibliography: see CFP for number of permitted pages

%\clearpage
\bibliographystyle{abbrv}  % do not change this line!
\bibliography{bibliography}  % put name of your .bib file here

\begin{thebibliography}{10}

\bibitem{Amato07:Optimizing}
C.~Amato, D.~S. Bernstein, and S.~Zilberstein.
\newblock Optimizing memory-bounded controllers for decentralized pomdps.
\newblock In {\em Twenty-Third Conference on Uncertainty in Artificial
  Intelligence (IJCAI)}, pages 1--8, 2007.

\bibitem{bampomdp}
C.~Amato and F.~A. Oliehoek.
\newblock Bayesian reinforcement learning for multiagent systems with state
  uncertainty.
\newblock In {\em Workshop on Multi-Agent Sequential Decision Making in
  Uncertain Domains}, pages 76--83, 2013.

\bibitem{amato}
C.~Amato and F.~A. Oliehoek.
\newblock Scalable planning and learning for multiagent pomdps.
\newblock In {\em Proceedings of the Twenty-Ninth AAAI Conference on Artificial
  Intelligence}, pages 1995--2002, 2015.

\bibitem{mcqalt}
B.~Banerjee, J.~Lyle, L.~Kraemer, and R.~Yellamraju.
\newblock Sample bounded distributed reinforcement learning for decentralized
  pomdps.
\newblock In {\em AAAI}, 2012.

\bibitem{mcesip}
R.~Ceren, P.~Doshi, and B.~Banerjee.
\newblock Reinforcement learning in partially observable multiagent settings:
  Monte carlo exploring policies with pac bounds.
\newblock In {\em Proceedings of the 2016 International Conference on
  Autonomous Agents \& Multiagent Systems}, pages 530--538, 2016.

\bibitem{chang2004all}
Y.-h. Chang, T.~Ho, and L.~P. Kaelbling.
\newblock All learning is local: Multi-agent learning in global reward games.
\newblock In {\em Advances in Neural Information Processing Systems}, pages
  807--814, 2004.

\bibitem{infpomdp}
F.~Doshi-Velez.
\newblock The infinite partially observable markov decision process.
\newblock In {\em Advances in neural information processing systems}, pages
  477--485, 2009.

\bibitem{PALO}
R.~Greiner.
\newblock Palo: A probabilistic hill-climbing algorithm.
\newblock {\em Artificial Intelligence}, 84(1):177--208, 1996.

\bibitem{factoredmdp}
C.~Guestrin, D.~Koller, and R.~Parr.
\newblock Multiagent planning with factored mdps.
\newblock In {\em NIPS}, volume~1, pages 1523--1530, 2001.

\bibitem{Hausknecht15:Deep}
M.~J. Hausknecht and P.~Stone.
\newblock Deep recurrent q-learning for partially observable mdps.
\newblock {\em CoRR}, abs/1507.06527, 2015.

\bibitem{alignment}
L.~Kraemer and B.~Banerjee.
\newblock Rehearsal based multi-agent reinforcment learning of decentralized
  plans.
\newblock In {\em W14-The 8th Workshop Multiagent Sequential Decision Making
  Under Uncertainty (MSDM 2013)}, page~24, 2013.

\bibitem{Kraemer16:Multi-agent}
L.~Kraemer and B.~Banerjee.
\newblock Multi-agent reinforcement learning as a rehearsal for decentralized
  planning.
\newblock {\em Neurocomputing}, 190:82--94, 2016.

\bibitem{liuopt}
B.~Liu, S.~Singh, R.~L. Lewis, and S.~Qin.
\newblock Optimal rewards in multiagent teams.
\newblock In {\em 2012 IEEE International Conference on Development and
  Learning and Epigenetic Robotics (ICDL)}, pages 1--8, 2012.

\bibitem{irpr}
M.~Liu, X.~Liao, and L.~Carin.
\newblock The infinite regionalized policy representation.
\newblock In {\em Proceedings of the 28th International Conference on Machine
  Learning (ICML-11)}, pages 769--776, 2011.

\bibitem{mpomdp}
J.~V. Messias, M.~Spaan, and P.~U. Lima.
\newblock Efficient offline communication policies for factored multiagent
  {POMDP}s.
\newblock In {\em Advances in Neural Information Processing Systems (NIPS)},
  pages 1917--1925, 2011.

\bibitem{tigerpomdp}
R.~Nair, M.~Tambe, M.~Yokoo, D.~Pynadath, and S.~Marsella.
\newblock Taming decentralized pomdps: Towards efficient policy computation for
  multiagent settings.
\newblock In {\em IJCAI}, pages 705--711, 2003.

\bibitem{perkins}
T.~J. Perkins.
\newblock Reinforcement learning for pomdps based on action values and
  stochastic optimization.
\newblock In {\em AAAI/IAAI}, pages 199--204, 2002.

\bibitem{rl}
R.~S. Sutton and A.~G. Barto.
\newblock {\em Introduction to reinforcement learning}.
\newblock MIT Press, 1998.

\end{thebibliography}

\end{document}